%% file: main.tex
\documentclass[10pt,twocolumn,letterpaper]{article}

\usepackage[pagenumbers]{cvpr} 

\usepackage{multirow}
\usepackage{graphicx}
\usepackage{booktabs}
\usepackage{amsmath}
\usepackage{textcomp}
\usepackage{colortbl}
\usepackage{bbding}

\input{preamble}
\definecolor{cvprblue}{rgb}{0.21,0.49,0.74}
\usepackage[pagebackref,breaklinks,colorlinks,allcolors=cvprblue]{hyperref}

\def\paperID{} 
\def\confName{CVPR}
\def\confYear{2026}

\title{Advancing Cancer Prognosis with Hierarchical Fusion of Genomic, Proteomic and Pathology Imaging Data from a Systems Biology Perspective}

\author{
    Junjie Zhou\textsuperscript{1}, \;
    Bao Xue\textsuperscript{1}, \;
    Meiling Wang\textsuperscript{2}, \;
    Wei Shao\textsuperscript{1}\thanks{Corresponding author}, \;
    Daoqiang Zhang\textsuperscript{1} \\
    {\small \textsuperscript{1}The College of Artificial Intelligence, Nanjing University of Aeronautics and Astronautics}\\
    {\small \textsuperscript{2}School of Computer Science, Nanjing University of Posts and Telecommunications}
    \vspace{-2mm}
}

\begin{document}
\maketitle
\input{sec/0_abstract}    
\input{sec/1_intro}
\input{sec/2_related}
\input{sec/3_method}

\input{sec/4_experiment}
\input{sec/5_conclusion}

\newpage
{
    \small
    \bibliographystyle{ieeenat_fullname}
    \bibliography{main}
}

\input{sec/X_suppl}

\end{document}

%% file: sec/0_abstract.tex
\begin{abstract}
To enhance the precision of cancer prognosis, recent research has increasingly focused on multimodal survival methods by integrating genomic data and histology images. However, current approaches overlook the fact that the proteome serves as an intermediate layer bridging genomic alterations and histopathological features while providing complementary biological information essential for survival prediction. This biological reality exposes another architectural limitation: existing integrative analysis studies fuse these heterogeneous data sources in a flat manner that fails to capture their inherent biological hierarchy. To address these limitations, we propose HFGPI, a hierarchical fusion framework that models the biological progression from genes to proteins to histology images from a systems biology perspective. Specifically, we introduce Molecular Tokenizer, a molecular encoding strategy that integrates identity embeddings with expression profiles to construct biologically informed representations for genes and proteins. We then develop Gene-Regulated Protein Fusion (GRPF), which employs graph-aware cross-attention with structure-preserving alignment to explicitly model gene-protein regulatory relationships and generate gene-regulated protein representations. Additionally, we propose Protein-Guided Hypergraph Learning (PGHL), which establishes associations between proteins and image patches, leveraging hypergraph convolution to capture higher-order protein-morphology relationships. The final features are progressively fused across hierarchical layers to achieve precise survival outcome prediction. Extensive experiments on five benchmark datasets demonstrate the superiority of HFGPI over state-of-the-art methods.
\end{abstract}

%% file: sec/1_intro.tex
\section{Introduction}
\label{sec:intro}
Accurate prediction of patient prognosis is fundamental to personalized cancer treatment, enabling risk stratification and therapeutic decision-making. Whole slide images (WSIs), as the gold standard for cancer diagnosis, provide rich phenotypic information about cellular organization and tissue architecture~\cite{pell2019use,yagi2005digital}. Given that the gigapixel scale of WSIs presents substantial computational challenges, multiple instance learning (MIL) approaches have been developed for patient-level survival prediction~\cite{li2021dual,lu2021data,shao2021transmil}. While histopathology captures tumor morphology and microenvironment organization, it lacks insights into underlying molecular mechanisms. Bulk transcriptomics, which quantifies tissue-level gene expression, reveals molecular subtypes and pathway dysregulation associated with patient survival~\cite{anaya2016oncolnc,dai2019whole}. This has motivated multimodal fusion approaches integrating genomics with histopathology through cross-modal attention~\cite{chen2021multimodal}, optimal transport~\cite{xu2023multimodal}, and prototype-based fusion~\cite{zhang2024prototypical}, consistently outperforming unimodal baselines across cancer types.

Despite these advances, current multimodal frameworks \textbf{suffer from a critical oversight:} they integrate genomic data and histology images while overlooking the proteome, which serves as an intermediate layer bridging genomic alterations and histopathological features~\cite{nussinov2019protein}, and provides crucial complementary biological information for survival prediction.  Existing works~\cite{chen2021multimodal,jaume2024modeling} rely on genomic signatures and pathway activities, which inherently cannot account for the post-transcriptional and translational regulatory mechanisms that ultimately determine cellular phenotypes and tissue architecture~\cite{aebersold2016mass,mani2022cancer}. Such limitations are evident in clinical contexts where protein-level alterations, rather than genomic changes alone, determine observable histopathological features. For example, in breast cancer, ERBB2 amplification drives HER2 protein overexpression, which determines membranous staining patterns in histopathology. Clinical decisions depend on HER2 protein status via immunohistochemistry rather than ERBB2 mRNA levels, because protein overexpression drives the membranous staining and cellular clustering visible in histopathology~\cite{vance2009genetic,wolff2023human}. This exemplifies how the proteome serves as an indispensable layer that cannot be fully captured by genomic signatures or pathways. Without this layer, existing methods miss the key biological entities that directly execute cellular functions and shape tissue architecture, and thereby influence survival outcomes. 




This biological reality \textbf{exposes another architectural limitation:} existing integrative analysis studies fuse these heterogeneous data sources in a flat manner that fails to capture their inherent biological hierarchy. While recent studies have explored various fusion strategies, most still treat modalities within the same hierarchical tier, without reflecting the layered dependencies of biological organization. For instance, CMTA~\cite{zhou2023cross} constructs parallel encoder-decoders with cross-modal attention to align gene and image representations, while PIBD~\cite{zhang2024prototypical} learns the prototypes for selecting discriminative information and obtains the disentangled features from WSIs and genomic data. However, from a systems biology perspective, biological information flows through a hierarchical cascade: genes encode instructions, proteins execute functions, and these functions manifest as tissue morphology~\cite{carroll2000endless}. Existing multimodal architectures fail to model the hierarchical dependencies along this ${gene}\rightarrow{protein}\rightarrow{phenotype}$ pathway, resulting in suboptimal integration that fails to capture the mechanistic pathways linking molecular aberrations to morphological outcomes. This raises a fundamental question: \emph{can we achieve more effective multimodal integration by explicitly modeling the hierarchical biological relationships that connect molecular alterations to tissue morphology?}

Furthermore, existing survival prediction methods \textbf{treat expression profiles as isolated numerical vectors}, neglecting the intrinsic biological properties of genes and proteins themselves. Gene expression values quantify activity levels but do not encode functional annotations, co-expression patterns, or regulatory relationships that are essential for understanding disease mechanisms. Similarly, protein abundance measurements alone fail to capture the functional roles and morphological manifestations that proteins exhibit in tissue architecture. This limitation prevents models from leveraging the rich biological knowledge embedded in molecular identities, restricting their ability to learn biologically meaningful representations.

To address these limitations, we introduce the proteome as the intermediate layer bridging genotype and phenotype, and propose HFGPI (Hierarchical Fusion of Genomic, Proteomic and Pathology Imaging Data), a hierarchical fusion framework that models the biological progression from genes to proteins to histology images from a systems biology perspective. First, we propose Molecular Tokenizer, a molecular encoding strategy that integrates identity embeddings (Gene2Vec~\cite{du2019gene2vec} for genes and LLM-generated functional descriptions for proteins) with expression profiles, enabling the model to jointly reason about intrinsic molecular properties and their patient-specific activities. Second, we develop Gene-Regulated Protein Fusion (GRPF), which employs graph-aware cross-attention with structure-preserving alignment to explicitly capture directional gene-to-protein regulatory influences. Third, we introduce Protein-Guided Hypergraph Learning (PGHL), which constructs biologically motivated hypergraphs linking proteins to their morphological manifestations, naturally modeling the many-to-many relationships between molecular functions and distributed tissue patterns. The final features are progressively fused across hierarchical layers to achieve precise survival outcome prediction. Extensive experiments demonstrate that HFGPI achieves state-of-the-art performance on five benchmark datasets while revealing interpretable cross-modal relationships among genomic, proteomic and pathology imaging data.



%% file: sec/2_related.tex
\section{Related Work}
\label{sec:related}
\subsection{Proteomics in Cancer Prognosis}
Proteomics has emerged as a critical layer of biological information in cancer research, as proteins are the primary executors of cellular functions and direct determinants of phenotypic outcomes~\cite{aebersold2016mass,mani2022cancer}. Unlike transcriptomics, which measures gene expression at the mRNA level, proteomics quantifies the actual functional molecules that drive cellular processes, tumor progression, and treatment response~\cite{aslam2016proteomics,liu2016dependency}. 
The biological importance of proteins in cancer prognosis stems from their direct influence on tissue morphology. This protein-morphology relationship is evident in widely used prognostic markers, for example, HER2 protein status correlates with specific membranous staining patterns and cellular architecture visible in routine histopathology~\cite{vance2009genetic,wolff2023human}; Ki-67 expression levels reflect proliferative activity through mitotic figures and cellular density patterns~\cite{scholzen2000ki,sun2018ki}; E-cadherin expression determines cellular cohesion, with loss resulting in discohesive growth patterns characteristic of invasive lobular carcinoma~\cite{canas2016cadherin}.

\subsection{Multimodal Survival Analysis}
Deep learning for cancer survival prediction has evolved from analyzing histopathology images in isolation to integrating multiple data modalities. Early works focus exclusively on WSIs, employing MIL methods to aggregate patch-level features into slide-level predictions~\cite{shao2021transmil,zhang2022dtfd,li2024dynamic}. 
Multimodal approaches emerge to integrate genomic data with histopathology for improved prognostic accuracy. 
For instance, MCAT~\cite{chen2021multimodal} introduces co-attention transformers that enable histology patches and genes to captures multimodal interactions when predicting survival outcomes. MOTCat~\cite{xu2023multimodal} employs optimal transport to establish soft correspondences between genomic data and image patches, providing global awareness to capture structural interactions within the tumor microenvironment. SurvPath~\cite{jaume2024modeling} incorporates biological pathway knowledge and applies multimodal Transformers to model interactions between pathway and histology patch tokens. LD-CVAE~\cite{zhou2025robust} and G-HANet~\cite{wang2025histo} are proposed to address the practical challenge of missing genomic data in clinical settings. More recently, PS3~\cite{raza2025ps3} extends beyond genomics and histology by integrating pathology reports as an additional textual modality. However, these approaches commonly neglect the proteomic layer that bridges genomic alterations and phenotypic manifestations, and they fuse heterogeneous data sources in a flat manner that fails to capture the biological cascade from genes to proteins to phenotypes. Consequently, such architectures struggle to represent the hierarchical dependencies inherent in biological systems, limiting their ability to enable precise survival predictions.

%% file: sec/3_method.tex
\section{Method}

\begin{figure*}[htbp]
  \centering
  \includegraphics[width=0.9\linewidth]{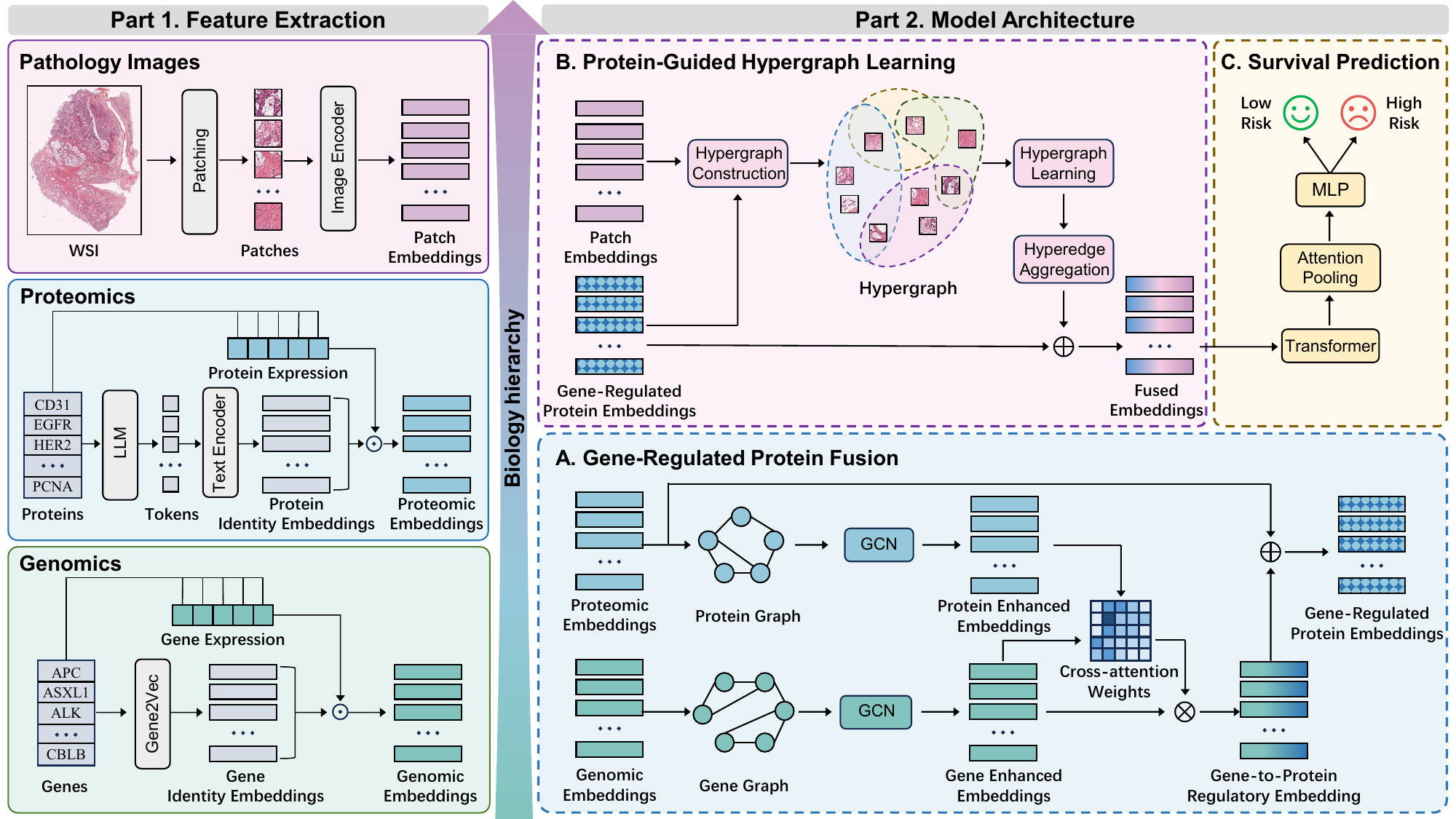}
  \caption{Overview of the HFGPI framework. The pipeline consists of two main components: (1) \textbf{Feature Extraction}: Whole slide histopathology images are partitioned into patches and encoded via a pre-trained vision backbone, while genomic and proteomic data are encoded by our Molecular Tokenizer that integrates identity embeddings with expression profiles. (2) \textbf{Model Architecture}: HFGPI models the biological hierarchy from genes to proteins to  histology images. Specifically, Gene-Regulated Protein Fusion (GRPF) employs graph-aware cross-attention with structure-preserving alignment to obtain the gene-regulated protein embeddings. Then, Protein-Guided Hypergraph Learning (PGHL) associates proteins with semantically related patches through hypergraph convolution, capturing higher-order protein-morphology relationships. Finally, the hierarchically fused features are aggregated via Transformer encoder and attention pooling to produce patient-level embeddings for survival prediction.}
  \label{fig:framework}
\end{figure*}

\subsection{Overview}
\label{sec:overview}
\cref{fig:framework} illustrates the overview of HFGPI. Given a patient's WSIs, genomic data and proteomic data, our framework proceeds through four stages that reflect the biological hierarchy from genes to proteins to tissue morphology. First, we extract patch-level features from WSIs using a pre-trained vision encoder, and encode molecular data i.e., genes and proteins, through our proposed Molecular Tokenizer, which integrates identity embeddings with expression profiles (\cref{sec:feature_extraction}). Second, we introduce Gene-Regulated Protein Fusion (GRPF) to model gene-protein regulatory relationships through graph-aware cross-attention with structure-preserving alignment, generating gene-regulated protein representations (\cref{sec:GRPF}). Third, Protein-Guided Hypergraph Learning (PGHL) constructs a hypergraph where each protein defines a hyperedge connecting semantically related image patches, and applies hypergraph convolution to capture higher-order protein-morphology relationships (\cref{sec:PGHL}). Finally, we aggregate the hierarchically fused features through a Transformer encoder and attention-based pooling for survival prediction (\cref{sec:survival_prediction}).

\subsection{Problem Formulation}
\label{sec:problem_formulation}

For the $k$-th patient, the available data is represented as  
$I^{(k)} = (\mathbf{Y}^{(k)}, \mathbf{X}_g^{(k)}, \mathbf{X}_p^{(k)}, c^{(k)}, t^{(k)})$, 
where $\mathbf{Y}^{(k)}$ denotes WSI patch features, 
$\mathbf{X}_g^{(k)}$ and $\mathbf{X}_p^{(k)}$ are genomic and proteomic embeddings, 
$c^{(k)} \in \{0,1\}$ indicates the right uncensorship status, and $t^{(k)} \in \mathbb{R}^+$ is the overall survival time. Then, we can estimate the hazard function $h^{(k)}(t|I^{(k)})=h^{(k)}(T=t|T \geq t, I^{(k)})$, which represents the probability of the event occurring instantaneously at time point $t$, and the survival function
$S^{(k)}(t) = \prod\nolimits_{u=1}^{t}\big(1 - h^{(k)}(u|I^{(k)})\big)$, which measures the likelihood of surviving beyond time $t$. Our model learns multimodal representations $f(I^{(k)})$ and is optimized by minimizing the negative log-likelihood (NLL) loss~\cite{zadeh2020bias}:
\begin{equation}
\begin{aligned}
\mathcal{L}_{\text{surv}}
&= -\frac{1}{N_D} \sum_{k=1}^{N_D} \Big[
 c^{(k)} \log (S^{(k)}(t|f(I^{(k)}))) \\
&+ (1 - c^{(k)}) \log (S^{(k)}(t-1|f(I^{(k)}))) \\
&+ (1 - c^{(k)}) \log (h^{(k)}(t|f(I^{(k)})))
\Big],
\label{eq:loss_surv}
\end{aligned}
\end{equation}
where $N_D$ is the number of patients in the training set.

\subsection{Feature Extraction}
\label{sec:feature_extraction}

\subsubsection{Histopathological Embeddings}
Given an input WSI $P^{(k)}$ for the $k$-th patient, we first segment the tissue regions and then partition the WSI into a set of non-overlapping patches at 20$\times$ magnification, denoted as $\{p^{(k)}_j\}_{j=1}^{M}$, where $M$ represents the total number of patches. A pre-trained image encoder $\varphi_{p}(\cdot)$ from vision-language models (VLMs) followed by a fully connected layer is employed to extract pathological features $Y^{(k)}$:
\begin{equation}
\mathbf{Y}^{(k)}=\{\varphi_{p}(p^{(k)}_j)\}_{j=1}^{M}=\{ y^{(k)}_1, y^{(k)}_2,..., y^{(k)}_{M} \}
\end{equation}

\subsubsection{Genomic Embeddings}
In previous survival prediction tasks~\cite{chen2021multimodal,jaume2024modeling}, gene expression profiles are typically treated as the sole genomic factor, as they directly quantify gene activity within biological systems. However, genes themselves possess intrinsic biological properties, including functional annotations and co-expression relationships, which are essential for understanding their contributions to disease progression. To capture both aspects, we propose \textbf{Molecular Tokenizer}, a molecular encoding strategy that integrates quantitative expression profiles with qualitative identity embeddings.

Following~\cite{yang2022scbert,xu2024multimodal}, we employ Gene2Vec~\cite{du2019gene2vec} to generate 200-dimensional identity embeddings $\mathbf{G} = [\mathbf{g}_1, \mathbf{g}_2, \ldots, \mathbf{g}_{N_g}]^T \in \mathbb{R}^{N_g \times d_g}$ that encode functional relationships and co-expression patterns, ensuring that functionally related genes are positioned proximally in the embedding space. For the $k$-th patient, given gene expression measurements $\mathbf{e}^{(k)} = [e_1^{(k)}, e_2^{(k)}, \ldots, e_{N_g}^{(k)}]^T \in \mathbb{R}^{N_g}$, we integrate identity embeddings with expression profiles as:
\begin{equation}
    \mathbf{X}_g^{(k)} = \mathbf{e}^{(k)} \odot \mathbf{G} \in \mathbb{R}^{N_g \times d_g},
\end{equation}
where $\odot$ denotes element-wise multiplication with broadcasting. This operation allows expression levels to modulate identity embeddings at the individual gene level, creating representations that encode both gene identity (what each gene is) and expression level (how much it is expressed).

\subsubsection{Proteomic Embeddings}
For proteomic data, we adopt the Molecular Tokenizer to obtain embeddings that capture both protein expression levels and biological semantics. Critically, we construct protein identity embeddings that naturally align with the histopathology image space through VLMs, facilitating subsequent protein-patch associations. Specifically, we leverage a large language model (LLM) to generate textual descriptions capturing: (1) the fundamental biological function of each protein, and (2) its potential correlations with morphological features observable in hematoxylin and eosin (HE) stained images. These generated descriptions are then encoded using the pre-trained text encoder from the VLM to obtain protein identity embeddings $\mathbf{P} = [\mathbf{p}_1, \mathbf{p}_2, \ldots, \mathbf{p}_{N_p}]^T \in \mathbb{R}^{N_p \times d}$. Given protein expression measurements $\mathbf{q}^{(k)} \in \mathbb{R}^{N_p}$ for the $k$-th patient, the proteomic embeddings are computed as:
\begin{equation}
    \mathbf{X}_p^{(k)} = \mathbf{q}^{(k)} \odot \mathbf{P} \in \mathbb{R}^{N_p \times d}.
\end{equation}
\textbf{Notation.} For clarity, we omit the patient index superscript $k$ 
in subsequent sections as the processing pipeline is identical for each patient sample.

\subsection{Gene-Regulated Protein Fusion}
\label{sec:GRPF}

Biological regulation flows directionally from genes to proteins through transcription and translation. To capture this hierarchical relationship, we propose Gene-Regulated Protein Fusion (GRPF), which models the regulatory process through graph-aware cross-attention with structure-preserving alignment.

\subsubsection{Graph-Aware Cross-Attention}

\noindent \textbf{Molecular Graph Construction.}
Genes and proteins do not function in isolation but form intricate regulatory and functional networks. Genes with similar expression patterns across conditions often participate in shared biological pathways (co-expression networks), while proteins with similar functional roles tend to physically interact or operate in coordinated complexes~\cite{allocco2004quantifying,jha2022prediction}. To encode these network relationships, we construct gene and protein graphs $\mathbf{A}_g$ and $\mathbf{A}_p$ from identity embeddings $\mathbf{G} \in \mathbb{R}^{N_g \times d_g}$ and $\mathbf{P} \in \mathbb{R}^{N_p \times d}$, respectively. We adopt the k-nearest neighbor (k-NN) algorithm to construct these graphs. For each graph, we define the neighborhood of each node as the top-$k_{i}$ ($i\in \{g,p\}$) most similar nodes based on cosine similarity, producing sparse adjacency matrices $\mathbf{A}_g$ and $\mathbf{A}_p$. 

\noindent \textbf{Graph-Aware Representation Refinement.}
To incorporate network context into molecular representations, we refine the genomic and proteomic embeddings $\mathbf{X}_g$ and $\mathbf{X}_p$ via Graph Convolutional Networks (GCNs)~\cite{kipf2016semi} that propagate information along functional connections:
\begin{equation}
    \mathbf{X}_g \leftarrow \text{GCN}(\mathbf{X}_g, \mathbf{A}_g), \quad \mathbf{X}_p \leftarrow \text{GCN}(\mathbf{X}_p, \mathbf{A}_p).
\end{equation}

\noindent \textbf{Cross-Attention for Gene-to-Protein Regulation.}
To model the directional gene-to-protein regulatory flow, we employ cross-attention where proteins query regulatory information from genes, reflecting the biological principle that genes control protein activity:

\begin{equation}
\begin{aligned}
    \mathbf{Q} &= \mathbf{X}_p \mathbf{W}_Q, \quad \mathbf{K} = \mathbf{X}_g \mathbf{W}_K, \quad \mathbf{V} = \mathbf{X}_g \mathbf{W}_V, \\
    \mathbf{T} &= \text{softmax}\left(\frac{\mathbf{Q}\mathbf{K}^T}{\sqrt{d}}\right) \in \mathbb{R}^{N_p \times N_g},
\end{aligned}
\end{equation}
where $\mathbf{W}_Q, \mathbf{W}_K, \mathbf{W}_V$ are learnable projection matrices. $\mathbf{T}$ represents the gene-to-protein co-attention matrix, with $T_{ij}$ representing the attention weight quantifying how much protein $i$ is regulated by gene $j$.

\subsubsection{Structure-Preserving Alignment}
\label{sec:structural_constraint}

In biological systems, functionally related proteins are often encoded by genes that exhibit coordinated transcriptional regulation~\cite{jansen2002relating,allocco2004quantifying}. This principle implies that genes encoding functionally coupled proteins should themselves be functionally related in the genomic regulatory network. To enforce this biological coherence, we introduce a structure-preserving alignment constraint~\cite{xu2019gromov} that encourages the learned attention matrix $\mathbf{T}$ to respect the intrinsic network topologies of both gene and protein modalities:
\begin{equation}
    \mathcal{L}_{\text{struct}} = \frac{1}{N_g N_p}\|\mathbf{C}_g - \mathbf{T}^T\mathbf{C}_p\mathbf{T}\|_F^2,
    \label{eq:loss_structure}
\end{equation}
where $\mathbf{C}_g = \mathbf{1} - \mathbf{A}_g$ and $\mathbf{C}_p = \mathbf{1} - \mathbf{A}_p$ are structural cost matrices, with low costs indicating high functional similarity.

\subsubsection{Gene-Regulated Protein Representation}

Finally, we obtain gene-regulated protein embeddings through cross-modal fusion:
\begin{equation}
    \mathbf{X}_p^{\text{regulated}} = \mathbf{X}_p + \mathbf{T}\mathbf{V} \in \mathbb{R}^{N_p \times d},
\end{equation}
where the first term preserves original protein information and the second term incorporates regulatory signals from genes.

\subsection{Protein-Guided Hypergraph Learning}
\label{sec:PGHL}
Proteins execute their biological functions through spatially distributed morphological alterations~\cite{uhlen2015tissue}. This manifests as a many-to-many relationship: a single protein may be expressed across multiple tissue regions, while individual image patches typically reflect the concurrent activity of multiple proteins~\cite{schapiro2017histocat}. To explicitly model these complex higher-order dependencies between proteins and tissue morphology, we propose Protein-Guided Hypergraph Learning (PGHL), which constructs biologically motivated hypergraphs linking proteins to their morphological manifestations.

\subsubsection{Protein-Patch Hypergraph Construction}
We model the protein-morphology relationship as a hypergraph $\mathcal{H} = (\mathcal{V}, \mathcal{E})$, where nodes $\mathcal{V}$ correspond to image patches and hyperedges $\mathcal{E}$ correspond to proteins. Each protein $i$ defines a hyperedge connecting the top-$k$ most relevant patches based on semantic similarity.

To identify relevant patches for each protein, we compute cosine similarity between patch features $\mathbf{Y}$ and gene-regulated protein embeddings $\mathbf{X}_p^{\text{regulated}}$:
\begin{equation}
    \mathbf{S} = \text{sim}(\mathbf{Y}, \mathbf{X}_p^{\text{regulated}}) \in \mathbb{R}^{M \times N_p}.
\end{equation}
Then, we select the top-$k$ patches with highest similarity scores to form a hyperedge. The hypergraph incidence matrix $\mathbf{H} \in \mathbb{R}^{M \times N_p}$ is constructed as:
\begin{equation}
    H_{ji} = \begin{cases}
        1 & \text{if patch } j \in \text{top-}k \text{ for protein } i \\
        0 & \text{otherwise}
    \end{cases}.
\end{equation}
\subsubsection{Hypergraph Learning}
To capture higher-order protein-morphology associations, we perform message propagation over the constructed hypergraph $\mathbf{H}$. This enables each image patch to aggregate contextual information from other patches that share common protein associations. Specifically, the initial patch features $\mathbf{Y}$ are updated by hypergraph convolution~\cite{feng2019hypergraph} to obtain updated patch representations $\mathbf{Z}$:
\begin{equation}
    \mathbf{Z} = 
    \sigma\!\left(
    \mathbf{D}_v^{-1/2}
    \mathbf{H}\mathbf{W}_e
    \mathbf{D}_e^{-1}
    \mathbf{H}^\top
    \mathbf{D}_v^{-1/2}
    \mathbf{Y}\mathbf{W}_p
    \right),
\end{equation}
where $\mathbf{D}_v$ and $\mathbf{D}_e$ are diagonal degree matrices for nodes and hyperedges, respectively, $\mathbf{W}_e$ and $\mathbf{W}_p$ are learnable weight matrices, and $\sigma(\cdot)$ denotes the activation function.

\subsubsection{Hyperedge Aggregation and Feature Fusion}
Following hypergraph convolution, we aggregate patch-level features to derive protein-driven morphological representations:
\begin{equation}
    \mathbf{E} = \mathbf{H}^T\mathbf{Z} / \text{deg}(\mathcal{E}) \in \mathbb{R}^{N_p \times d},
\end{equation}
where $\text{deg}(\mathcal{E})$ denotes the degree of each hyperedge.

To obtain unified multimodal representations, we further fuse these protein-driven morphological features with gene-regulated protein embeddings:
\begin{equation}
    \mathbf{F} = \mathbf{E} + \mathbf{X}_p^{\text{regulated}},
\end{equation}
yielding hybrid representations that jointly encode gene regulation, protein semantics, and tissue morphology.

\subsection{Survival Prediction and Training Objective}
\label{sec:survival_prediction}
The fused feature representations $\mathbf{F}$ are first processed by a Transformer encoder $\mathcal{T}$ to capture global contextual dependencies and interactions among the integrated multimodal features. The refined embeddings $\mathbf{F}' = \mathcal{T}(\mathbf{F})$ are then aggregated into a compact patient-level representation through a gated attention pooling module~\cite{chen2021multimodal,lu2021data}, which adaptively weights each feature according to its prognostic relevance. The aggregated representation $\mathbf{h}$ is finally passed through a prediction head to estimate the hazard function $h^{(k)}(t|\mathbf{h})$.

\subsubsection{Training Objective}
The overall objective combines the survival loss in \cref{eq:loss_surv} $\mathcal{L}_{\text{surv}}$ with the structure-preserving alignment constraint $\mathcal{L}_{\text{struct}}$ introduced in \cref{eq:loss_structure}:
\begin{equation}
    \mathcal{L} = \mathcal{L}_{\text{surv}} + \lambda \mathcal{L}_{\text{struct}},
\end{equation}
where $\lambda$ controls the trade-off between predictive performance and structural coherence.

%% file: sec/4_experiment.tex
\input{tables/main_exp}

\section{Experiments}

\subsection{Datasets and Evaluation Metrics}

We use five cancer datasets from The Cancer Genome Atlas (TCGA) to evaluate HFGPI: Bladder Urothelial Carcinoma (BLCA) ($n=340$), Breast Invasive Carcinoma (BRCA) ($n=833$), Glioblastoma and Lower Grade Glioma (GBMLGG) ($n=878$), Lung Adenocarcinoma (LUAD) ($n=368$), and Uterine Corpus Endometrial Carcinoma (UCEC) ($n=122$). These cohorts are selected based on the availability of matched genomic, proteomic and pathology imaging data, enabling multimodal analysis. 

\noindent \textbf{Genomic Data.} Gene expression profiles are obtained from the UCSC Xena database~\cite{goldman2020visualizing}, originally generated using the Illumina HiSeq 2000 RNA-Seq platform. Expression values are RSEM normalized and log$_2$(x+1) transformed to stabilize variance and reduce the impact of outliers. We retain genes with sufficient expression variability across samples for downstream analysis.

\noindent \textbf{Proteomic Data.} Protein expression data are acquired from Reverse Phase Protein Array (RPPA) measurements curated by UCSC Xena~\cite{goldman2020visualizing}. These datasets provide unit-normalized expression values for cancer-related proteins and phosphoproteins, quantifying functional proteomic states that serve as intermediate phenotypes between genomic alterations and histopathological manifestations.

\noindent \textbf{Pathology Images.} The diagnostic WSIs are collected from the National Cancer Institute (NCI) Genomic Data Commons (GDC)~\cite{weinstein2013cancer}.

\noindent \textbf{Evaluation Metric.} We employ 5-fold cross-validation with the 4:1 ratio of training and validation sets. The model performance is measured by the mean concordance index (C-index)~\cite{harrell1996multivariable} with its standard deviation (std).


\subsection{Implementation Details}
\label{sec:implementation}

We employ Gene2Vec~\cite{du2019gene2vec} to extract gene identity embeddings and use an MLP as the encoder for genomic features. For proteomic data, we leverage GPT-5 to generate semantic descriptions of protein functions, which are then encoded using the CONCH~\cite{lu2024visual} text encoder. For histopathology images, we utilize the pre-trained CONCH image encoder to obtain patch-level features that naturally align with the protein feature space through shared architecture. We select the top $N_g = 2,000$ highly variable genes for analysis. For graph construction, we employ k-NN graphs with $k_g = 100$ for genes and $k_p = 20$ for proteins. The hypergraph learning module identifies the top-$k=32$ most relevant patches for each protein to construct hyperedges.

The model is trained for 20 epochs using the AdamW optimizer with an initial learning rate of $1 \times 10^{-4}$ and weight decay of $1 \times 10^{-5}$. We set the batch size to 1 with gradient accumulation over 16 steps. The structure-preserving regularization weight is $\lambda = 0.3$. All experiments are conducted on NVIDIA RTX 3090 GPUs. Additional implementation details are provided in the \textbf{Supplementary Material}.

\subsection{Comparisons with State-of-the-Art Methods}
\label{sec:baselines}

To demonstrate the effectiveness of our proposed method, we compare it against state-of-the-art unimodal and multimodal survival prediction models. For genomic and proteomic data, we employ Self-Normalizing Network (SNN)~\cite{klambauer2017self} to predict survival outcomes from the original gene or protein expression profiles. For histopathology, we compare with state-of-the-art MIL methods, including ABMIL~\cite{ilse2018attention}, CLAM~\cite{lu2021data}, TransMIL~\cite{shao2021transmil}, and WiKG~\cite{li2024dynamic}.
For multimodal integration of histopathological images with genomic data, we compare with several recent proposed methods such as MCAT~\cite{chen2021multimodal}, MOTCat~\cite{xu2023multimodal}, CMTA~\cite{zhou2023cross}, SurvPath~\cite{jaume2024modeling}, MMP~\cite{song2024multimodal}, PIBD~\cite{zhang2024prototypical}, and MoME~\cite{xiong2024mome}.
In addition, we compare with methods capable of handling three modalities, including PS3~\cite{raza2025ps3} and ICFNet~\cite{zhang2025icfnet}, where we adapt them by replacing their original text modality with proteomic data to align with our experimental setup. We summarize the key findings below.


\noindent \textbf{Unimodal vs. Multimodal.}
\cref{tab:main_results} presents performance comparisons across five TCGA cohorts. Unimodal approaches achieve limited prognostic results, with genomic/proteomic methods averaging C-index of 0.677 and 0.647, while histopathology-only methods reach 0.686. In contrast, multimodal integration consistently outperforms unimodal baselines. Our HFGPI achieves an average C-index of 0.753, significantly surpassing the best unimodal method i.e., WiKG, by 6.7\%. This substantial gain suggests that heterogeneous modalities capture complementary biological aspects essential for accurate survival prediction.

\noindent \textbf{Protein vs. No Protein.}
To evaluate the contribution of proteomic data, we compare three-modality (Gene + Protein + Image) with two-modality (Gene + Image) methods in \cref{tab:main_results}. Three-modality methods which incorporates proteomic information consistently outperform their two-modality counterparts, with improvements ranging from 1.1\% to 5.4\%. These results empirically demonstrate the advantages that the proteome serves as an essential intermediate phenotype, bridging the genotype-phenotype gap and offering complementary biological insights critical for precise survival prediction.

\noindent \textbf{Hierarchical vs. Flat Multimodal Fusion.}
HFGPI achieves state-of-the-art performance across all datasets, with notable improvements on GBMLGG and UCEC. Compared to three-modality methods, HFGPI outperforms PS3 by 1.8\% and ICFNet by 2.4\% on average. We attribute this superiority to our hierarchical fusion strategy, which explicitly models the information flow from genes to proteins to histology. In contrast, the comparing methods employ flat fusion that treats 
all modalities equally, without capturing the inherent biological dependencies among them. This biologically informed design enables HFGPI to capture coherent multimodal representations, leading to improved prognostic accuracy.

To further validate the prognostic stratification capability of our model, we provide Kaplan-Meier~\cite{kaplan1958nonparametric} survival curves and Log-rank test~\cite{mantel1966evaluation} statistics in the \textbf{Supplementary Material}.




\input{tables/main_ablation}

\subsection{Ablation Study}

\noindent \textbf{Impact of Each Modality.}
To dissect the contribution of each modality within our hierarchical framework, we systematically evaluate HFGPI variants by removing individual modalities. As shown in \cref{table:main_ablation}, removing any single modality consistently degrades performance. The full three-modality HFGPI consistently outperforms all two-modality variants on the average C-index, validating that genomic, proteomic, and histological features capture complementary biological aspects essential for accurate survival prediction.

\noindent \textbf{Impact of Molecular Tokenizer.}
To validate our molecular encoding strategy, we compare Molecular Tokenizer with alternative gene representations used in existing methods, such as gene families~\cite{chen2021multimodal}) and pathways~\cite{jaume2024modeling}. As shown in \cref{table:main_ablation}, using gene families reduces average C-index by 1.4\%, and pathway based encoding decreases performance by 1.0\%. These results demonstrate the superiority of our Molecular Tokenizer that encodes individual genes by integrating identity embeddings with expression values, preserving fine-grained molecular information. 

\noindent \textbf{Impact of Gene-Regulated Protein Fusion.}
We assess the contribution of GRPF by replacing it with standard cross-attention. As shown in \cref{table:main_ablation}, removing GRPF results in a 2.3\% performance drop. 
This degradation highlights the importance of GRPF for accurately modeling gene-protein regulatory relationships.

\noindent \textbf{Impact of Structure-Preserving Alignment.}
We propose a variant of HFGPI without $\mathcal{L}_{\text{struct}}$ to assess the contribution of the structure-preserving alignment. As shown in \cref{table:main_ablation}, removing $\mathcal{L}_{\text{struct}}$ leads to a 1.6\% performance degradation. This validates the the effectiveness of maintaining biological coherence through structure-preserving alignment.


\noindent \textbf{Impact of Protein-Guided Hypergraph Learning.}
To validate the effectiveness of PGHL in capturing protein-histology associations, we replace PGHL with cross-attention between protein embeddings and image patches. As shown in \cref{table:main_ablation}, replacing PGHL with standard cross-attention results in a 1.8\% performance drop. This degradation demonstrates the importance of modeling higher-order relationships in protein-morphology associations.

\input{tables/main_discussion}

\vspace{-3mm}
\subsection{Discussion}
\noindent \textbf{Impact of Different LLMs.}
To assess the robustness of our framework to LLMs for generating protein descriptions, we compare different LLMs including GPT-5, DeepSeek~\cite{liu2024deepseek}, Claude-3.7 and Qwen-3~\cite{yang2025qwen3}. As shown in \cref{tab:main_discussion}(a), GPT-5 achieves the best performance with 0.753 average C-index, followed by  Qwen-3 (0.747), Claude-3.7 (0.746) and DeepSeek (0.743). While GPT-5 slightly outperforms other models, all LLMs yield comparable results, indicating that our framework is robust to the choice of LLMs.

\noindent \textbf{Impact of Different VLMs.}
We further evaluate our framework with different VLMs for encoding both protein descriptions (text encoder) and histology images (image encoder). Specifically, we compare CONCH with CLIP~\cite{radford2021learning}, PLIP~\cite{huang2023visual} and QUILT~\cite{ikezogwo2023quilt}. \cref{tab:main_discussion}(b) shows that CONCH achieves the best performance, substantially outperforming the second-best VLM PLIP by 2.8\%. 
This can be attributed to the fact that CONCH is a stronger foundation model pre-trained on large-scale pathological image-caption pairs.

Additional discussions of parameters can be found in the \textbf{Supplementary Material}.

\begin{figure}[ht]
    \centering
    \includegraphics[width=0.95\linewidth]{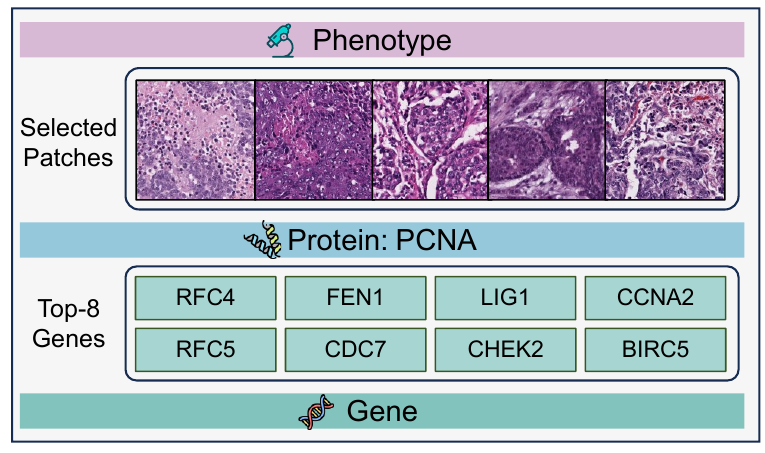}
    \caption{Visualization results on BRCA dataset using PCNA as an exemplar protein.}
    \label{fig:case_brca}
    \vspace{-3mm}
\end{figure}

\subsection{Visualization}
To demonstrate the hierarchical interpretability of HFGPI, we visualize the regulatory cascade from genes to proteins to tissue phenotypes using PCNA as an exemplar protein in the BRCA dataset, as shown in \cref{fig:case_brca}. PCNA (Proliferating Cell Nuclear Antigen) is a core component of the DNA replication machinery and serves as a marker of proliferative activity in breast cancer~\cite{jurikova2016ki67,malkas2006cancer}. To explore its upstream gene regulation, we compute the mean attention weights between PCNA and genes from the GRPF module across all samples in the BRCA dataset and identify the top-ranked regulatory genes, including \emph{RFC4}, \emph{FEN1}, \emph{LIG1}, and \emph{CCNA2}. These genes are involved in DNA replication and cell cycle control, which is consistent with PCNA's role in replication and tumor cell proliferation~\cite{kang2019regulation,zheng2011fen1}. At the protein-to-phenotype level, \cref{fig:case_brca} highlights representative image patches identified through PGHL module. These patches exhibit typical histopathological features associated with elevated PCNA expression, including increased nuclear-to-cytoplasmic ratio and dense tumor cellularity, reflecting the proliferative phenotype characteristic of aggressive breast cancer subtypes. Additional case studies are provided in the \textbf{Supplementary Material}.


%% file: tables/main_exp.tex
\begin{table*}[htbp]
\setlength{\arrayrulewidth}{0.1mm}  
\small 
\centering
\caption{Comparisons of C-index (mean ± std) with SOTA methods over five cancer datasets. G., P. and I. refer to genomic, proteomic and pathology imaging data, respectively. The best results and the second-best results are highlighted with \textbf{bold} and in \underline{underline}, respectively. Methods marked with $^{\dagger}$ are variants that replace the original textual data with proteomic data.}
\label{tab:main_results}
\renewcommand{\arraystretch}{1.1}
\resizebox{0.93\linewidth}{!}{
\begin{tabular}{c ccc cccccc}
\hline \hline
\textbf{Model}  & G. & P. & I. & \textbf{BLCA} & \textbf{BRCA} & \textbf{GBMLGG} & \textbf{LUAD} & \textbf{UCEC} & \textbf{Overall} \\
\hline
\hline
Gene exp.~\cite{klambauer2017self} & \checkmark & &  & 0.672$\pm$0.059 & 0.635$\pm$0.072 & 0.805$\pm$0.050 & 0.627$\pm$0.043 & 0.645$\pm$0.098 & 0.677 \\
\hline
Protein exp.~\cite{klambauer2017self} &  & \checkmark &  & 0.648$\pm$0.057 & 0.666$\pm$0.112 & 0.754$\pm$0.090 & 0.585$\pm$0.056 & 0.580$\pm$0.133 & 0.647 \\
\hline
ABMIL~\cite{ilse2018attention} & & & \checkmark & 0.657$\pm$0.044 & 0.698$\pm$0.061 & 0.788$\pm$0.065 & 0.602$\pm$0.034 & 0.582$\pm$0.112 & 0.665 \\
CLAM-SB~\cite{lu2021data} & & & \checkmark & 0.672$\pm$0.027 & 0.679$\pm$0.061 & 0.790$\pm$0.090 & 0.612$\pm$0.036 & 0.585$\pm$0.186 & 0.668 \\
CLAM-MB~\cite{lu2021data} & & & \checkmark & 0.682$\pm$0.038 & 0.675$\pm$0.086 & 0.790$\pm$0.087 & 0.632$\pm$0.012 & 0.594$\pm$0.123 & 0.675 \\
TransMIL~\cite{shao2021transmil} & & & \checkmark & 0.688$\pm$0.043 & 0.691$\pm$0.049 & 0.814$\pm$0.078 & 0.618$\pm$0.052 & 0.586$\pm$0.115 & 0.679 \\
WiKG~\cite{li2024dynamic} & & & \checkmark & 0.691$\pm$0.042 & 0.699$\pm$0.075 & 0.808$\pm$0.083 & 0.601$\pm$0.045 & 0.631$\pm$0.117 & 0.686 \\
\hline
MCAT~\cite{chen2021multimodal} & \checkmark & & \checkmark & 0.686$\pm$0.022 & 0.685$\pm$0.069 & 0.835$\pm$0.113 & 0.639$\pm$0.033 & 0.716$\pm$0.101 & 0.712 \\
MOTCat~\cite{xu2023multimodal} & \checkmark & & \checkmark & 0.688$\pm$0.031 & 0.686$\pm$0.045 & 0.827$\pm$0.143 & 0.623$\pm$0.011 & 0.705$\pm$0.060 & 0.706 \\
CMTA~\cite{zhou2023cross} & \checkmark & & \checkmark & 0.693$\pm$0.025 & 0.681$\pm$0.050 & 0.839$\pm$0.134 & 0.643$\pm$0.035 & 0.702$\pm$0.145 & 0.712 \\
SurvPath~\cite{jaume2024modeling} & \checkmark & & \checkmark & 0.685$\pm$0.042 & 0.694$\pm$0.054 & 0.817$\pm$0.177 & 0.648$\pm$0.052 & 0.688$\pm$0.130 & 0.706 \\
$\text{MMP}_{\text{Trans.}}$~\cite{song2024multimodal} & \checkmark & & \checkmark & 0.689$\pm$0.015 & 0.671$\pm$0.078 & 0.815$\pm$0.085 & 0.639$\pm$0.038 & 0.690$\pm$0.150 & 0.701 \\
PIBD~\cite{zhang2024prototypical} & \checkmark & & \checkmark & 0.679$\pm$0.059 & 0.689$\pm$0.040 & 0.802$\pm$0.092 & 0.624$\pm$0.048 & 0.702$\pm$0.100 & 0.699 \\
MoME~\cite{xiong2024mome} & \checkmark & & \checkmark & 0.704$\pm$0.048 & 0.688$\pm$0.066 & 0.835$\pm$0.031 & 0.651$\pm$0.023 & 0.714$\pm$0.120 & 0.718 \\
\hline
PS3$^{\dagger}$~\cite{raza2025ps3} & \checkmark & \checkmark & \checkmark & \underline{0.708$\pm$0.039} & \underline{0.702$\pm$0.074} & \underline{0.851$\pm$0.099} & 0.659$\pm$0.038 & \underline{0.757$\pm$0.120} & \underline{0.735} \\
ICFNet$^{\dagger}$~\cite{zhang2025icfnet} & \checkmark & \checkmark & \checkmark & 0.705$\pm$0.025 & 0.692$\pm$0.061 & 0.846$\pm$0.139 & \underline{0.664$\pm$0.035} & 0.739$\pm$0.089 & 0.729 \\
\rowcolor{gray!20} Ours & \checkmark & \checkmark & \checkmark & \textbf{0.717$\pm$0.022} & \textbf{0.715$\pm$0.043} & \textbf{0.873$\pm$0.064} & \textbf{0.680$\pm$0.039} & \textbf{0.782$\pm$0.062} & \textbf{0.753} \\
\hline \hline
\end{tabular}
}
\end{table*}

%% file: tables/main_ablation.tex
\begin{table}[ht]
\small
    \centering
    \caption{Ablation study on key components, with C-index averaged across five datasets.}
    \label{table:main_ablation}
    \resizebox{0.95\linewidth}{!}{
    \begin{tabular}{c|c|c}
    \toprule
    \textbf{Ablation} & \textbf{Model} & \textbf{Avg.} \\ 
    \midrule
    \textbf{Full Model} & \textbf{HFGPI}& \textbf{0.753} \\ 
    \midrule
    \multirow{3}{*}{Modalities} 
    & $g,p,i$ $\Rightarrow$ $g,p$  & $0.708$ ($-4.5\%$) \\ 
    & $g,p,i$ $\Rightarrow$ $p,i$  & $0.708$ ($-4.5\%$) \\ 
    & $g,p,i$ $\Rightarrow$ $g,i$  & $0.713$ ($-4.0\%$) \\ 
    \midrule
    \multirow{2}{*}{Tokenizer}
    & Molecular. $\Rightarrow$ Families  & $0.739$ ($-1.4\%$) \\ 
    & Molecular. $\Rightarrow$ Pathways  & $0.743$ ($-1.0\%$) \\ 
    \midrule
    \multirow{2}{*}{Fusion}
    & GRPF $\Rightarrow$ CA & $0.730$ ($-2.3\%$) \\ 
    & PGHL $\Rightarrow$ CA & $0.735$ ($-1.8\%$) \\ 
    \midrule
    Struct. & w/ $\mathcal{L}_{struct}$ $\Rightarrow$ w/o $\mathcal{L}_{struct}$ & $0.737$ ($-1.6\%$) \\
    \bottomrule
    \end{tabular}
    }
    \vspace{-3mm}
\end{table}

%% file: tables/main_discussion.tex
\begin{table}[htbp]
    \centering
    \caption{Performance of using different (a) VLMs and (b) LLMs on average C-index across five datasets.}
    \label{tab:main_discussion} 
    \begin{subtable}{.48\linewidth}
      \centering
        \caption{Different VLMs.} 
        \label{tab:avg_cindex_vlm}
        \resizebox{!}{1.0cm}{%
        \begin{tabular}{l|c}
            \hline
            \textbf{Model} & \textbf{Avg.} \\
            \hline
            CONCH & \textbf{0.753} \\
            CLIP  & $0.688$ ($-6.5\%$) \\
            PLIP  & $0.725$ ($-2.8\%$) \\
            QUILT & $0.716$ ($-3.7\%$) \\
            \hline
        \end{tabular}%
        }
    \end{subtable}%
    \hfill
    \begin{subtable}{.48\linewidth}
      \centering
        \caption{Different LLMs.} 
        \label{tab:avg_cindex_llm}
        \resizebox{!}{1.0cm}{%
        \begin{tabular}{l|c}
            \hline
            \textbf{Model} & \textbf{Avg.} \\
            \hline
            GPT-5      & \textbf{0.753} \\
            DeepSeek   & $0.743$ ($-1.0\%$) \\
            Qwen-3       & $0.747$ ($-0.6\%$) \\
            Claude-3.7 & $0.746$ ($-0.7\%$) \\
            \hline
        \end{tabular}%
        }
    \end{subtable}
\end{table}

%% file: sec/5_conclusion.tex
\section{Conclusion}

In this work, we present HFGPI, a hierarchical fusion framework that models the biological progression from
genes to proteins to histology images from a systems biology perspective. Specifically, we introduce Molecular Tokenizer to encode individual genes and proteins with both identity and expression profiles. Subsequently, we develop Gene-Regulated Protein Fusion (GRPF) to model directional gene-to-protein regulatory relationships, and propose Protein-Guided Hypergraph Learning (PGHL) to capture higher-order protein-morphology associations. By incorporating proteomics as an intermediate layer bridging genotype and phenotype, HFGPI achieves state-of-the-art performance on five TCGA datasets.

%% file: sec/X_suppl.tex
\clearpage
\setcounter{page}{1}
\maketitlesupplementary

\section{Survival Analysis}
\label{app:survival}
\subsection{Hazard and Survival Functions}
For the $k$-th patient with data $I^{(k)} = (\mathbf{Y}^{(k)}, \mathbf{X}_g^{(k)}, \mathbf{X}_p^{(k)}, c^{(k)}, t^{(k)})$, 
where $\mathbf{Y}^{(k)}$ denotes WSI patch features, $\mathbf{X}_g^{(k)}$ and $\mathbf{X}_p^{(k)}$ represent genomic and proteomic embeddings, $c^{(k)} \in \{0,1\}$ indicates the censorship status, and $t^{(k)} \in \mathbb{R}^+$ is the survival time in months.

The hazard function $h^{(k)}(t|I^{(k)})$ quantifies the instantaneous risk of death at time $t$, defined as:
\begin{equation}
h^{(k)}(t|I^{(k)}) = \lim_{\Delta t \to 0} \frac{P(t \leq T < t+\Delta t | T \geq t, I^{(k)})}{\Delta t}.
\end{equation}

The survival function $S^{(k)}(t|I^{(k)})$ represents the probability of surviving beyond time $t$, which can be expressed in terms of the cumulative hazard:
\begin{equation}
\begin{aligned}
S^{(k)}(t|I^{(k)}) &= \prod\nolimits_{u=1}^{t}\big(1-h^{(k)}(u|I^{(k)})\big)\\ 
&= \exp\left(-\sum_{u=1}^{t} h^{(k)}(u|I^{(k)})\right).
\end{aligned}
\end{equation}

\subsection{Cox Proportional Hazards Model}
The Cox Proportional Hazards (CoxPH) model is a widely used approach for estimating the hazard function, in which $h(t|I^{(k)})$ is parameterized as:
\begin{equation}
h(t|I^{(k)}) = h_{0}(t) \exp(\theta^\top f(I^{(k)})),
\end{equation}
where $h_{0}(t)$ is the baseline hazard function, $\theta$ is a vector of coefficients, and $f(I^{(k)})$ represents the learned multimodal representation.

\subsection{Negative Log-Likelihood Loss}
Following~\cite{zadeh2020bias}, we optimize our model using the negative log-likelihood (NLL) loss for discrete-time survival analysis:
\begin{equation}
\begin{aligned}
\mathcal{L}_{\text{surv}}
&= -\frac{1}{N_D} \sum_{k=1}^{N_D} \Big[
 c^{(k)} \log S^{(k)}(t^{(k)}|f(I^{(k)})) \\
&+ (1 - c^{(k)}) \log S^{(k)}(t^{(k)}-1|f(I^{(k)})) \\
&+ (1 - c^{(k)}) \log h^{(k)}(t^{(k)}|f(I^{(k)}))
\Big],
\end{aligned}
\end{equation}
where the first term accounts for observed events (uncensored cases), and the second and third terms handle censored observations by maximizing the probability of surviving up to the censoring time while penalizing the hazard at that time.

\section{Additional Detailed Implementation}
\label{sec:supp_implementation}

\subsection{Molecular Feature Extraction}

\noindent\textbf{Genomic Data Processing.}
For genomic data, we employ Gene2Vec~\cite{du2019gene2vec} to extract 200-dimensional gene identity embeddings that capture functional relationships between genes. Given the high dimensionality of gene expression profiles (approximately 30,000 genes per sample), we perform feature selection by identifying the top $N_g = 2,000$ genes with the highest variance across the entire cohort. This variance-based selection ensures that we retain the most informative genes that exhibit significant biological variability relevant to patient outcomes.

The gene expression values are first normalized using log-transformation: $\tilde{x}_i = \log_2(x_i + 1)$, where $x_i$ represents the raw expression count for gene $i$. Subsequently, we apply z-score normalization across samples to standardize the expression profiles:
\begin{equation}
x_i^{\text{norm}} = \frac{\tilde{x}_i - \mu_i}{\sigma_i}
\end{equation}
where $\mu_i$ and $\sigma_i$ denote the mean and standard deviation of gene $i$ across all samples.

\noindent\textbf{Proteomic Data Processing.} For proteomic data, we leverage GPT-5 to generate comprehensive semantic descriptions of each protein's biological function, subcellular localization, and morphological correlates in tissue histology. The prompt template used for description generation is:

\begin{quote}
\textit{``Describe the biological function and typical histopathological appearance of [protein name] in cancer tissue. Include information about its visual characteristics that would be observable in H\&E stained tissue sections.''}
\end{quote}

\noindent These generated descriptions are then encoded using the CONCH~\cite{lu2024visual} text encoder, which produces 512-dimensional protein identity embeddings. Crucially, this encoding naturally aligns the protein feature space with the histopathology image space, as CONCH is trained contrastively on paired pathology images and text descriptions.

\medskip
\noindent\textbf{Example Protein Descriptions:}

\begin{itemize}[leftmargin=*, itemsep=6pt]
    \item \textbf{4E-BP1:} A translational repressor protein that regulates mRNA translation by binding eIF4E. Its elevated expression is often associated with increased cellular proliferation and altered nuclear morphology detectable in H\&E-stained tissue sections.
    
    \item \textbf{Akt:} A serine/threonine kinase involved in regulating cell survival, growth, and metabolism. Elevated Akt expression often correlates with increased cellular density and atypical morphology observable in H\&E-stained tissue sections.
    
    \item \textbf{Bax:} A pro-apoptotic member of the Bcl-2 protein family that promotes programmed cell death. Higher Bax expression typically correlates with increased cellular shrinkage, nuclear condensation, and tissue architecture disruption observable in H\&E-stained sections.
    
    \item \textbf{CD31:} An endothelial cell adhesion molecule (PECAM-1) that facilitates leukocyte transmigration. Higher expression correlates with increased vascular density and prominent endothelial structures visible in H\&E-stained tissue sections.
    
    \item \textbf{CDK1:} A serine/threonine kinase that regulates cell cycle progression. Its elevated expression is often associated with increased mitotic activity and higher nuclear atypia observable in H\&E-stained tissue sections.
\end{itemize}

\begin{figure*}[htbp]
    \centering
    \includegraphics[width=1.0\linewidth]{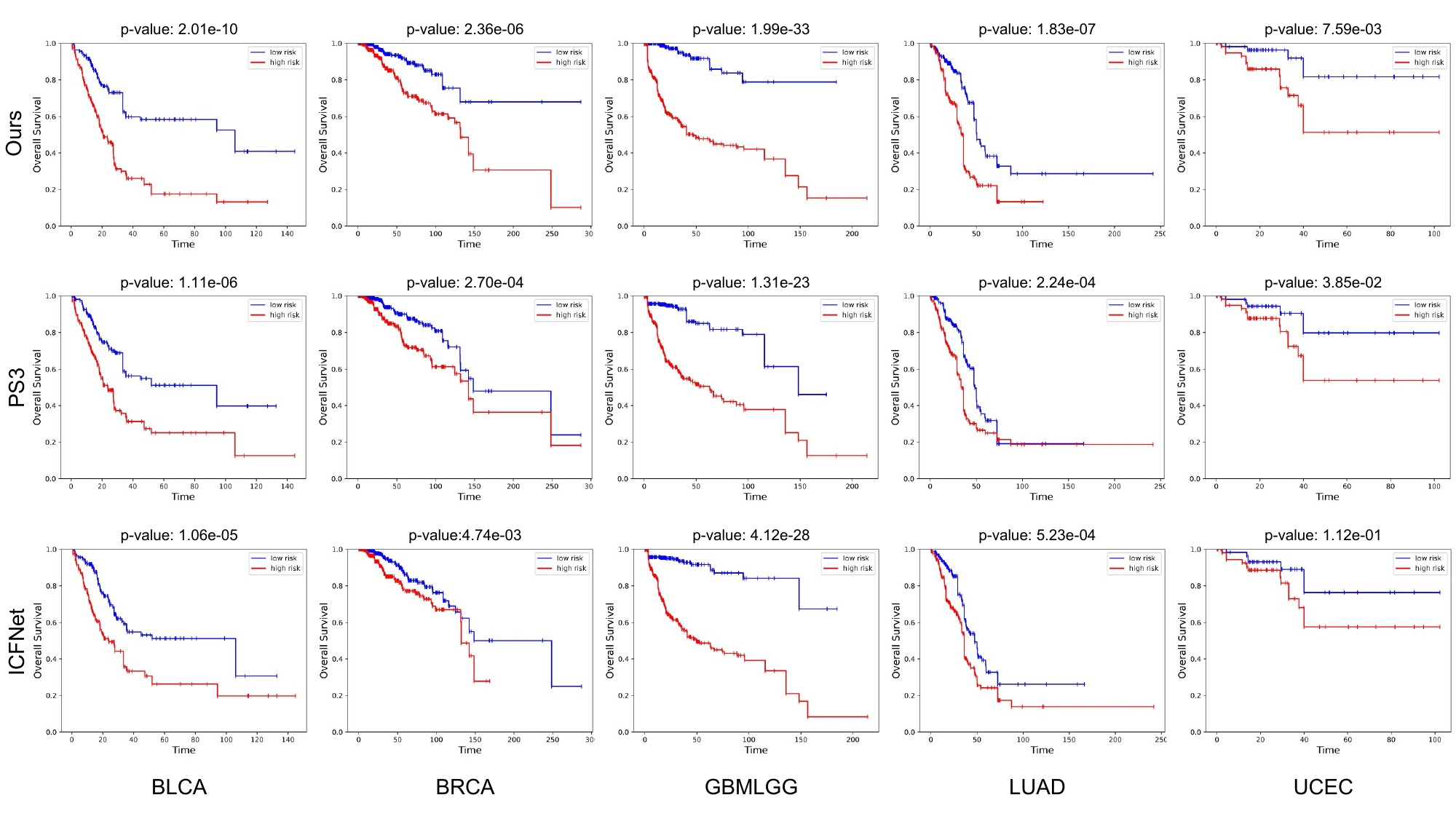}
    \caption{Kaplan-Meier Analysis of predicted high-risk (red) and low-risk (blue) groups on five cancer datasets.}
    \label{fig:survival_curves}
\end{figure*}

\input{tables/table_genenum}

\input{tables/table_kg}

\input{tables/table_kp}

\input{tables/table_topk}

\input{tables/table_lambda}

\subsection{Histopathological Feature Extraction}

\noindent\textbf{Whole Slide Image Processing.}
All whole slide images (WSIs) are processed at 20$\times$ magnification, corresponding to a resolution of approximately 0.5 $\mu$m/pixel. We apply an automated tissue segmentation pipeline to distinguish tissue regions from background.

\noindent\textbf{Patch Extraction and Encoding.}
The remaining tissue regions are cropped into non-overlapping patches of size 512$\times$512 pixels. For each WSI, this typically yields between 1,000 and 10,000 patches depending on tissue size. We employ the pre-trained CONCH~\cite{lu2024visual} image encoder to extract 512-dimensional patch-level features. The CONCH encoder ensures semantic alignment with protein embeddings through its shared architecture and contrastive pre-training on pathology-specific image-text pairs.

\section{Baseline Methods}

\subsection{Uni-modal Baselines}

\noindent \textbf{SNN~\cite{klambauer2017self}:} Self-Normalizing Networks employ scaled exponential linear units (SELUs) to enable self-normalization properties, which we apply to gene and protein expression profiles for survival prediction.

\noindent \textbf{ABMIL~\cite{ilse2018attention}:} ABMIL uses an attention mechanism to aggregate patch-level features from whole slide images, enabling the model to focus on diagnostically relevant regions for survival prediction.

\noindent \textbf{CLAM~\cite{lu2021data}:} CLAM incorporates instance-level clustering to identify distinct morphological patterns and uses attention-based aggregation for slide-level survival prediction.

\noindent \textbf{TransMIL~\cite{shao2021transmil}:} TransMIL employs self-attention mechanisms to capture long-range dependencies among image patches, enabling more effective context-aware feature aggregation for histopathology-based survival analysis.

\noindent \textbf{WiKG~\cite{li2024dynamic}:} WiKG constructs dynamic instance graphs to model spatial relationships among tissue regions and employs graph neural networks for survival prediction from histopathological images.

\subsection{Two-modality Baselines}

\noindent \textbf{MCAT~\cite{chen2021multimodal}:} MCAT employs cross-modal attention to learn interpretable, dense co-attention mappings between WSI features and genomic profiles within a shared embedding space.

\noindent \textbf{MOTCat~\cite{xu2023multimodal}:} MOTCat employs optimal transport to establish soft correspondences between genomic data and image patches, providing global awareness to capture structural interactions within the tumor microenvironment.

\noindent \textbf{CMTA~\cite{zhou2023cross}:} CMTA introduces two parallel encoder-decoder structures to integrate intra-modal information and generate cross-modal representation.

\noindent \textbf{SurvPath~\cite{jaume2024modeling}:} SurvPath proposes a memory-efficient, resolution-agnostic multimodal Transformer that integrates transcriptomic pathway tokens and histology patch tokens for patient survival prediction.

\noindent \textbf{MMP~\cite{song2024multimodal}:} MMP learns prototypical representations for each modality, with the resulting multimodal tokens processed by a fusion network using either Transformers or optimal transport-based cross-alignment.

\noindent \textbf{PIBD~\cite{zhang2024prototypical}:} PIBD designs a Prototypical Information Bottleneck (PIB) that models prototypes for selecting discriminative information to reduce intra-modal redundancy, while Prototypical Information Disentanglement (PID) addresses inter-modal redundancy by decoupling multimodal data into distinct components with the guidance of joint prototypical distribution.

\noindent \textbf{MoME~\cite{xiong2024mome}:} MoME designs a mixture of multimodal experts layer which enables the network to selectively focus on the information from a specific modality and utilizes the reference information in different forms across encoding stages.

\subsection{Three-modality Baselines}

\noindent \textbf{PS3~\cite{raza2025ps3}:} PS3 proposes a prototype-based multimodal fusion framework that integrates cancer aggressiveness-related signals from whole slide images, pathology reports, and transcriptomic data for improved survival prediction.

\noindent \textbf{ICFNet~\cite{zhang2025icfnet}:} ICFNet integrates histopathology whole slide images, genomic expression profiles, and clinical textual data. It employs three distinct encoders to extract modality-specific features and leverages optimal transport algorithms to align and fuse interrelated features across modalities.

To enable fair comparison on our three-modality setting (genomic, proteomic and pathology imaging data), we maintain the original architectures of PS3 and ICFNet but replace their textual modality (pathology reports or clinical text) with proteomic features.

\subsection{Patient Stratification}
Beyond prediction accuracy, effective patient stratification into distinct risk groups is crucial for personalized treatment planning. We evaluate stratification performance using Kaplan-Meier survival analysis, where patients are divided into high-risk and low-risk groups based on median predicted risk scores, with statistical significance assessed via the log-rank test.
As presented in \cref{fig:survival_curves}, our approach demonstrates significantly improved discrimination between the two groups and achieves the lowest p-values when compared to baseline methods. 
These results demonstrate that our hierarchical fusion of genomic, proteomic, and pathology imaging data yields biologically meaningful risk stratification with strong statistical power, essential for clinical decision-making.

\subsection{Additional Visualization Results}

\begin{figure}[ht]
    \centering
    \includegraphics[width=1.0\linewidth]{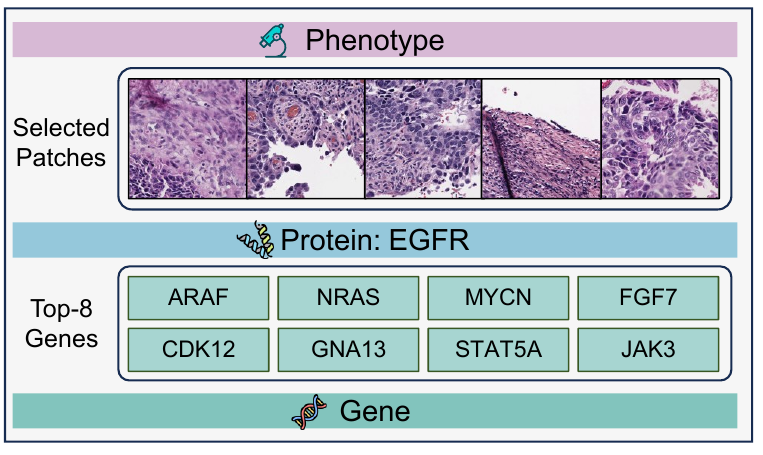}
    \caption{Visualization results on BLCA dataset using EGFR as an exemplar protein.}
    \label{fig:case_blca}
\end{figure}

\begin{figure}[ht]
    \centering
    \includegraphics[width=1.0\linewidth]{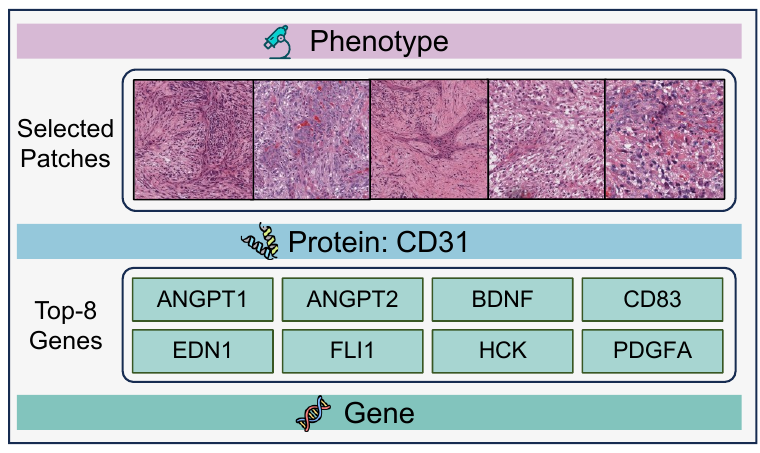}
    \caption{Visualization results on GBMLGG dataset using CD31 as an exemplar protein.}
    \label{fig:case_gbmlgg}
\end{figure}

To further demonstrate the hierarchical interpretability of HFGPI, we provide additional case studies in the BLCA (EGFR protein) and GBMLGG (CD31 protein) datasets, as shown in \cref{fig:case_blca,fig:case_gbmlgg}.

\textbf{Case Study 1: EGFR in BLCA Dataset.}
EGFR (Epidermal Growth Factor Receptor) is a receptor tyrosine kinase that is frequently overexpressed in bladder cancer and plays a critical role in tumorigenesis through activation of downstream signaling pathways~\cite{cheng2002overexpression,chakravarti2005expression,rose2020egfr}. In \cref{fig:case_blca}, we identify genes associated with EGFR, including \emph{ARAF}, \emph{NRAS}, \emph{FGF7} and \emph{STAT5A}. These genes are involved in key signaling pathways that promote cell growth and tumor progression, which aligns with EGFR's role in driving cancer cell proliferation~\cite{su2022araf,therkildsen2014predictive,ferguson2021fibroblast}. At the protein-to-phenotype level, representative image patches from the PGHL module exhibit typical aggressive tumor features, including high nuclear-to-cytoplasmic ratio, prominent nucleoli, nuclear pleomorphism, solid tumor architecture with loss of normal urothelial differentiation, and increased mitotic activity. These histopathological patterns reflect the proliferative and invasive phenotype characteristic of EGFR-driven bladder cancer.

\textbf{Case Study 2: CD31 in GBMLGG Dataset.}
CD31 (PECAM-1, Platelet Endothelial Cell Adhesion Molecule-1) is a transmembrane glycoprotein predominantly expressed on endothelial cells and serves as a critical marker for tumor-associated vasculature. Microvascular proliferation is a histopathological hallmark of glioblastomas, and CD31 expression correlates with tumor neovascularization, disease progression, and patient prognosis~\cite{tamura2018histopathological,moghaddam2015evaluation}. In \cref{fig:case_gbmlgg}, we identify genes associated with CD31, including \emph{ANGPT1}, \emph{ANGPT2}, \emph{FLT1}, and \emph{PDGFB}. These genes are involved in angiogenesis and vascular remodeling, which aligns with CD31's role in mediating endothelial cell interactions and tumor neovascularization~\cite{chen2024pericytes,chappell2016flt,hellstrom1999role}. Representative image patches from the PGHL module display characteristic features of glioblastoma vasculature, including microvascular proliferation with glomeruloid tufts, multilayered endothelial cell networks, and regions of palisading necrosis surrounded by hypercellular areas. These morphological patterns are consistent with the aggressive angiogenic phenotype associated with CD31 expression in high-grade gliomas.


\section{Additional Discussions}

\subsection{Number of Selected Genes}

Highly variable genes capture the most informative transcriptional variations across samples. We investigate the impact of gene selection by varying $N_g \in \{1000, 2000, 3000\}$. As shown in \cref{tab:genenum}, the model achieves optimal performance with $N_g = 2000$. Using fewer genes ($N_g = 1000$) leads to information loss, reducing the C-index to 0.742, while including more genes ($N_g = 3000$) introduces noisy features that do not contribute to survival prediction and slightly degrades performance. These results suggest that $N_g = 2000$ provides an appropriate balance between capturing biologically relevant variations and avoiding noise.

\subsection{Number of Neighbors in Gene Graph}
We evaluate neighborhood sizes $k_g \in \{50, 100, 150\}$ for graph construction. As shown in \cref{tab:kg}, $k_g = 100$ yields the best performance. When $k_g = 50$, the graph is too sparse to capture sufficient functional relationships, while $k_g = 150$ introduces excessive connections that include biologically irrelevant interactions. Thus, $k_g = 100$ provides an optimal balance between capturing meaningful gene interactions and maintaining graph sparsity.

\subsection{Number of Neighbors in Protein Graph}
We similarly evaluate $k_p \in \{10, 20, 30\}$ for protein graph construction. \cref{tab:kp} shows that $k_p = 20$ achieves optimal performance, with smaller values ($k_p = 10$) providing insufficient connectivity and larger values ($k_p = 30$) introducing weakly related proteins.

\subsection{Number of Top-$k$ Patches in PGHL}
The PGHL module constructs hyperedges by selecting the top-$k$ most relevant patches for each protein. We investigate the effect of hyperedge size with $k \in \{8, 16, 32, 64, 128\}$. As shown in \cref{tab:topk}, $k = 32$ achieves the best performance. Smaller values ($k \leq 16$) provide insufficient spatial coverage of protein-related morphological patterns, while larger values ($k \geq 64$) incorporate patches with weak protein-morphology associations, introducing noise. Thus, $k = 32$ optimally balances coverage of protein-relevant patches with association strength.

\subsection{Effect of $\lambda$}
\cref{tab:lambda} summarizes the influence of the structure-preserving regularization on model performance. Incorporating the regularization can improve the overall performance, with $\lambda = 0.3$ achieving the best balance between structural coherence and predictive accuracy. These results indicate that insufficient regularization fails to constrain the attention map, leading to noisy gene–protein associations, whereas excessive regularization forces the learned matrix $\mathbf{T}$ to overfit the network topology, thereby weakening the survival prediction.

\input{tables/table_computation}

\subsection{Computational Complexity}
To evaluate the practical efficiency of HFGPI, we compare its computational cost with state-of-the-art three-modality methods on the BRCA dataset. As shown in \cref{tab:complexity}, HFGPI achieves superior efficiency compared to ICFNet while remaining comparable to PS3. Specifically, HFGPI requires only 2.03~M parameters with an inference time of 0.017~s/slide and training time of 2228~s. In contrast, ICFNet demands 6.18~M parameters with significantly higher computational costs: 0.104~s/slide for inference and 11232~s for training. These results demonstrate that HFGPI achieves superior prognostic performance while maintaining high computational efficiency, making it practical for clinical deployment.


%% file: tables/table_genenum.tex
\begin{table*}[htbp]
\setlength{\arrayrulewidth}{0.1mm}  
\tiny
\centering
\caption{Impact of the number of selected genes ($N_g$) on model performance across five TCGA datasets.}
\label{tab:genenum}
\resizebox{0.95\linewidth}{!}{
\begin{tabular}{c cccccc}
\hline
\hline
\textbf{$N_{g}$} & \textbf{BLCA} & \textbf{BRCA} & \textbf{GBMLGG} & \textbf{LUAD} & \textbf{UCEC} & \textbf{Overall} \\
\hline
\hline
1000 & 0.711$\pm$0.033 & 0.709$\pm$0.049 & 0.862$\pm$0.056 & 0.650$\pm$0.042 & 0.779$\pm$0.074 & 0.742 \\
\rowcolor{gray!15} 2000 & 0.717$\pm$0.022 & 0.715$\pm$0.043 & 0.873$\pm$0.064 & 0.680$\pm$0.039 & 0.782$\pm$0.062 & 0.753 \\
3000 & 0.710$\pm$0.066 & 0.698$\pm$0.038 & 0.847$\pm$0.071 & 0.662$\pm$0.042 & 0.751$\pm$0.071 & 0.734 \\
\hline
\hline
\end{tabular}
}
\end{table*}

%% file: tables/table_kg.tex
\begin{table*}[htbp]
\setlength{\arrayrulewidth}{0.1mm}  
\tiny 
\centering
\caption{Impact of the number of neighbors ($k_g$) in gene graph construction on model performance across five TCGA datasets.}
\label{tab:kg}
\resizebox{0.95\linewidth}{!}{
\begin{tabular}{c cccccc}
\hline
\hline
$k_{g}$ & \textbf{BLCA} & \textbf{BRCA} & \textbf{GBMLGG} & \textbf{LUAD} & \textbf{UCEC} & \textbf{Overall} \\
\hline
\hline
50 & 0.703$\pm$0.037 & 0.715$\pm$0.027 & 0.864$\pm$0.087 & 0.674$\pm$0.041 & 0.778$\pm$0.062 & 0.747 \\
\rowcolor{gray!15} 100 & 0.717$\pm$0.022 & 0.715$\pm$0.043 & 0.873$\pm$0.064 & 0.680$\pm$0.039 & 0.782$\pm$0.062 & 0.753 \\
150 & 0.711$\pm$0.055 & 0.705$\pm$0.069 & 0.865$\pm$0.090 & 0.668$\pm$0.051 & 0.763$\pm$0.073 & 0.742 \\
\hline
\hline
\end{tabular}
}
\end{table*}

%% file: tables/table_kp.tex
\begin{table*}[htbp]
\setlength{\arrayrulewidth}{0.1mm}  
\tiny 
\centering
\caption{Impact of the number of neighbors ($k_p$) in protein graph construction on model performance across five TCGA datasets.}
\label{tab:kp}
\resizebox{0.95\linewidth}{!}{
\begin{tabular}{c cccccc}
\hline
\hline
\textbf{$k_{p}$} & \textbf{BLCA} & \textbf{BRCA} & \textbf{GBMLGG} & \textbf{LUAD} & \textbf{UCEC} & \textbf{Overall} \\
\hline
\hline
10 & 0.698$\pm$0.047 & 0.694$\pm$0.050 & 0.842$\pm$0.131 & 0.670$\pm$0.065 & 0.762$\pm$0.081 & 0.733 \\
\rowcolor{gray!15} 20 & 0.717$\pm$0.022 & 0.715$\pm$0.043 & 0.873$\pm$0.064 & 0.680$\pm$0.039 & 0.782$\pm$0.062 & 0.753 \\
30 & 0.699$\pm$0.042 & 0.717$\pm$0.037 & 0.854$\pm$0.103 & 0.673$\pm$0.034 & 0.781$\pm$0.031 & 0.745 \\
\hline
\hline
\end{tabular}
}
\end{table*}

%% file: tables/table_topk.tex
\begin{table*}[htbp]
\setlength{\arrayrulewidth}{0.1mm}  
\tiny 
\centering
\caption{Impact of the number of top-$k$ patches selected per protein in PGHL on model performance across five TCGA datasets.}
\label{tab:topk}
\resizebox{0.95\linewidth}{!}{
\begin{tabular}{c cccccc}
\hline
\hline
\textbf{$k$} & \textbf{BLCA} & \textbf{BRCA} & \textbf{GBMLGG} & \textbf{LUAD} & \textbf{UCEC} & \textbf{Overall} \\
\hline
\hline
8 & 0.694$\pm$0.031 & 0.684$\pm$0.039 & 0.871$\pm$0.047 & 0.665$\pm$0.038 & 0.730$\pm$0.069 & 0.729 \\
16 & 0.707$\pm$0.061 & 0.697$\pm$0.051 & 0.878$\pm$0.049 & 0.672$\pm$0.048 & 0.770$\pm$0.024 & 0.745 \\
\rowcolor{gray!15} 32 & 0.717$\pm$0.022 & 0.715$\pm$0.043 & 0.873$\pm$0.064 & 0.680$\pm$0.039 & 0.782$\pm$0.062 & 0.753 \\
64 & 0.707$\pm$0.053 & 0.708$\pm$0.041 & 0.866$\pm$0.080 & 0.678$\pm$0.025 & 0.775$\pm$0.122 & 0.747 \\
128 & 0.703$\pm$0.055 & 0.704$\pm$0.051 & 0.864$\pm$0.071 & 0.683$\pm$0.024 & 0.760$\pm$0.110 & 0.743 \\

\hline
\hline
\end{tabular}
}
\end{table*}

%% file: tables/table_lambda.tex
\begin{table*}[htbp]
\setlength{\arrayrulewidth}{0.1mm}  
\small
\centering
\caption{Impact of the regularization parameter ($\lambda$) on overall model performance (average C-index) across five TCGA datasets.}
\label{tab:lambda}
\resizebox{0.7\linewidth}{!}{
\begin{tabular}{c|ccccccc}
\hline
{Metric} & {0} & {0.1} & {0.3} & {0.5} & {0.7} & {0.9} & {1.0} \\
\hline
{Average C-index} 
& 0.738 
& 0.741 
& \textbf{0.753} 
& 0.747 
& 0.741 
& 0.735 
& 0.730 \\
\hline
\end{tabular}
}
\end{table*}

%% file: tables/table_computation.tex

\begin{table}[htp]
\centering
\caption{Comparisons of model complexity and efficiency. We report the number of parameters (MB), inference time per slide (ms), and training time (s) on the BRCA dataset.}
\label{tab:complexity}
\resizebox{1.\linewidth}{!}{
\begin{tabular}{lccc}
\hline
\hline
\multirow{2}{*}{\textbf{Methods}} & \textbf{\#Params} & \textbf{Inference Time} & \textbf{Training Time} \\
&  {(M)} & {(s / slide)} & {(s)} \\
\hline
PS3 & 1.08 & 0.013 & 1876 \\
ICFNet & 6.18 & 0.104 & 11232 \\
\rowcolor{gray!15}
\textbf{Ours} & 2.03 & 0.017 & 2228 \\
\hline
\hline
\end{tabular}
}
\end{table}